\documentclass{article}

\usepackage{arxiv}

\usepackage[utf8]{inputenc} 
\usepackage[T1]{fontenc}    
\usepackage{hyperref}       
\usepackage{url}            
\usepackage{booktabs}       
\usepackage{amsfonts}       
\usepackage{nicefrac}       
\usepackage{microtype}      
\usepackage{lipsum}
\usepackage{graphicx}
\graphicspath{ {./images/} }
\usepackage{tabularx}
\usepackage{wrapfig}
\usepackage{wasysym}

\newcolumntype{Y}{>{\centering\arraybackslash}X}

\usepackage{xcolor}
\usepackage{soul}

\definecolor{axe}{rgb}{0.921568627,0.294117647,0.266666667}
\definecolor{chicken}{rgb}{0.984313725,0.921568627,0.317647059}
\definecolor{hammer}{rgb}{0.619607843,0.988235294,0.333333333}
\definecolor{horse}{rgb}{.46,.98,.30}
\definecolor{owl}{rgb}{0.474509804,0.984313725,0.576470588}
\definecolor{seagull}{rgb}{0.458823529,0.984313725,0.701960784}
\definecolor{zebra}{rgb}{0.2509803,0.349019608,0.964705882}

\definecolor{hand_wave}{rgb}{0.984313725,0.921568627,0.317647059}
\definecolor{scratch_head}{rgb}{0.474509804,0.984313725,0.576470588}
\definecolor{take_photo}{rgb}{0.392156863,0.843137255,0.984313725}
\definecolor{throw_catch}{rgb}{0.317647059,0.690196078,0.976470588}

\title{Grounded Language Acquisition From Object and Action Imagery}

\author{
 James R. Kubricht \\
  GE Aerospace\\
  Niskayuna, NY 12309 \\
  \texttt{james.kubricht@ge.com} \\
   \And
 Zhaoyuan Yang \\
  GE Vernova \\
  Niskayuna, NY 12309 \\
  \texttt{zhaoyuan.yang@ge.com} \\
  \And
 Jianwei Qiu \\
  GE HealthCare \\
  Niskayuna, NY 12309 \\
  \texttt{jianwei.qiu@ge.com} \\
  \And
 Peter H. Tu \\
  GE Aerospace\\
  Niskayuna, NY 12309 \\
  \texttt{tu@ge.com} \\
}

\begin{document}
\maketitle
\begin{abstract}
Deep learning approaches to natural language processing have made great strides in recent years. While these models produce words (symbols) that convey vast amounts of diverse knowledge, their output is not grounded in any explicit form of sensory data or corresponding knowledge. In this paper, we explore the development of a private language for visual data representation by training emergent language (EL) encoder/decoder networks in both i) a traditional referential game environment and ii) a contrastive learning environment utilizing a within-class matching training paradigm. An additional classification layer--utilizing neural machine translation and random forest classification--was used to transform symbolic representations (sequences of integer symbols) to class labels. These methods were applied in two experiments focusing on object recognition and action recognition. For object recognition, a set of sketches produced by human participants from real imagery was used (Sketchy dataset) and for action recognition, 2D trajectories were generated from 3D motion capture systems (MoVi dataset). In order to interpret the symbols produced for data in each experiment, Gradient-weighted Class Activation Mapping (Grad-CAM) methods were used to identify pixel regions indicating semantic features which contribute evidence towards symbols in learned languages. Additionally, a t-distributed Stochastic Neighbor Embedding (t-SNE) method was used to investigate embeddings learned by CNN feature extractors. 
\end{abstract}


\section{Introduction}

Recent advancements in natural language processing (NLP) have produced remarkable results, generating models which provide language output that exceeds many human intellectual capacities. These large language models (LLMs)~\cite{zhao2023survey} have been touted as a major step in the pursuit of artificial general intelligence (AGI), motivated in part by the finding that architecture scaling enables the emergence of fascinating capabilities. LLMs, however, may succumb to erroneous language predictions under certain conditions, e.g., hallucination in open-domain problems~\cite{ouyang2022training}. In more ideal cases, embeddings from symbolic inputs are processed into language outputs that are indistinguishable from those produced by humans. Hallucinations, therefore, occur when symbols are embedded and manipulated in ways that do not accord with ontological constraints in either the world or the specified domain. This is reminiscent of Searle's \cite{searle1980minds} Chinese room thought experiment, wherein a person efficiently processes Chinese characters but produces symbols that are not grounded in mental models~\cite{gentner2014mental} borne from their experience.

With regards to the pursuit of AGI, it is therefore critical to develop AI systems which are capable of manipulating symbols through slow and deliberate (System 2) reasoning processes~\cite{kahneman2011thinking} that are ultimately grounded in sensory experience. With this objective in mind, the current work aims to demonstrate how machines can learn to communicate through a learned, private language based on embeddings from convolutional neural network (CNN) feature extractors. The experiments conducted in this paper build on prior work exploring the utility of emergent languages~\cite{bouchacourt2018agents, chaabouni2020compositionality, havrylov2017emergence, lazaridou2020emergent} in constructing explainable, symbolic representations of deep learning architectures. Purposes of these architectures have included: i) segmentation and classification in medical imagery~\cite{chowdhury2020escell, chowdhury2020symbolic, chowdhury2021emergent, santamaria2020towards}; ii) symbolic VAEs~\cite{devaraj2020symbols}; iii) semantic action analysis~\cite{kubricht2020emergent, santamaria2019towards}; and iv) automated language acquisition~\cite{kubricht2022towards}.

This paper builds on prior work by pursuing novel methods for explaining and exploring symbols in learned languages with respect to visual semantic features that enable higher-level reasoning processes\footnote{This research was, in part, funded by the U.S. Government. The views and conclusions contained in this document are those of the authors and should not be interpreted as representing the official policies, either expressed or implied, of the U.S. Government. Use, duplication, or disclosure is subject to the restrictions as stated in Agreement HR00111990061 between the Government and GE.}. An overview of the architecture used to encode and decode input imagery is provided in Section~\ref{sec:emergent_language} followed by two approaches to language learning: referential game training (Section~\ref{sec:referential}) and contrastive learning~\cite{schroff2015facenet} (Section~\ref{sec:contrastive}). Methods for predicting class labels using symbols are described in Section~\ref{sec:classification} followed by an overview of visualization methods for explainability (Section~\ref{sec:visualization}) and the utilized datasets (Section~\ref{sec:datasets}). Results for each dataset are provided in Section~\ref{sec:experiments} and conclusions on are made in Section~\ref{sec:conclusion}.

\section{Technical Description}

\begin{figure}[h!]
  \includegraphics[width=\textwidth]{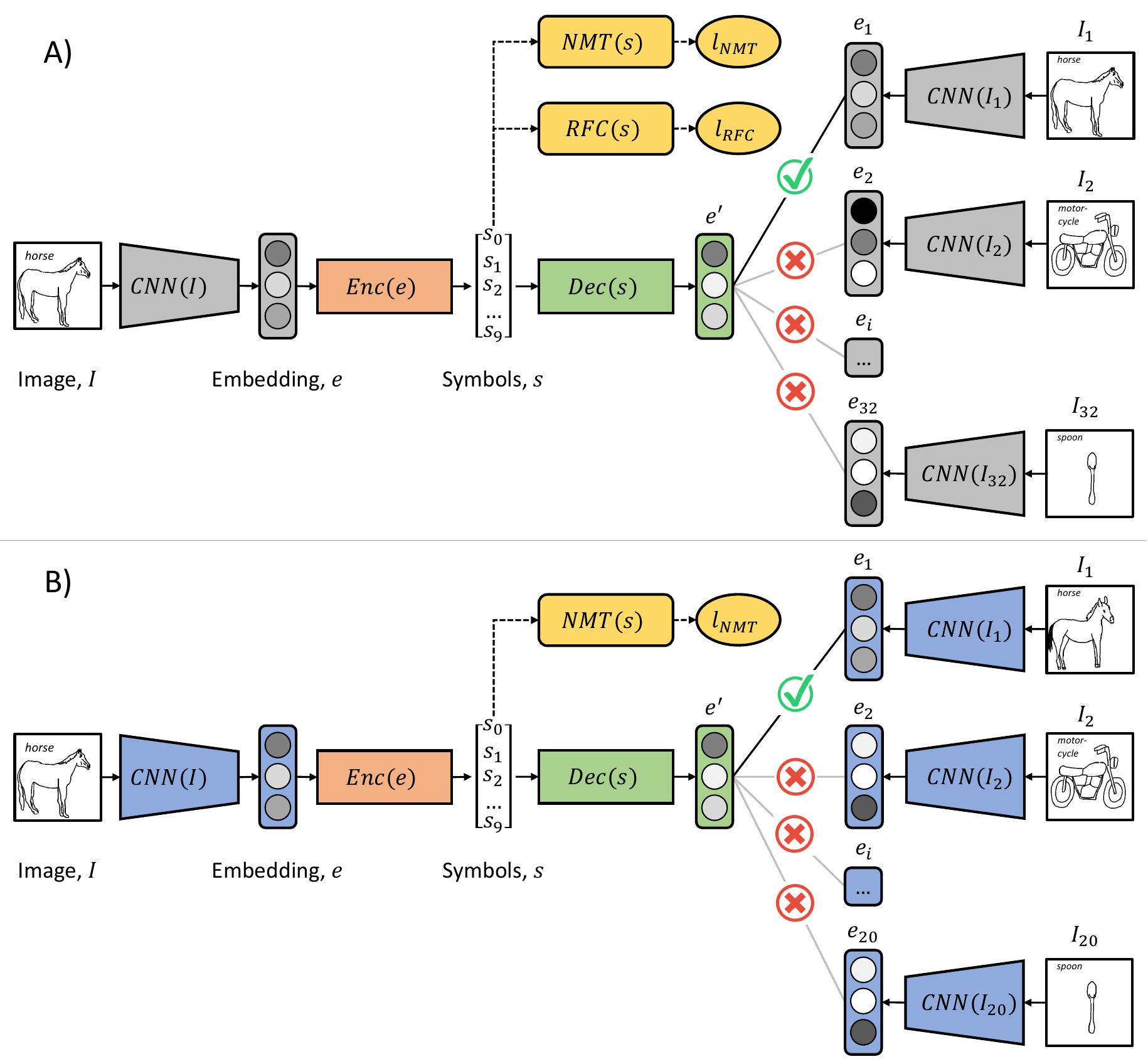}
  \vspace{-6mm}
  \caption{Diagrams outlining systems developed in Experiments 1 and 2 using referential game and contrastive loss learning approaches, respectively. A) In Experiment 1, an image $I$ is transformed into an embedding $e$ using a pretrained convolutional neural network, $CNN$. The embedding is encoded into a set of symbols, $s$, and decoded into a reconstructed embedding, $e'$. A referential game is then played, where $e'$ is matched with its true embedding among a set of 31 distractors, $e_{i:i\neq0}$. B) In Experiment 2, a convolutional neural network is trained to produce an embedding which is encoded and decoded into an anchor embedding, $e_A$. The anchor is matched to a different image belonging to the same class (positive example, $e_P$) amongst an image belonging to a different class (negative example; $e_N$). Symbols $s$ are classified into class labels $l_{NMT}$ and $l_{RFC}$ using neural machine translation $NMT$ and random forest classification $RFC$, respectively.}
  \vspace{-2mm}
  \label{fig:system_diagram}
\end{figure}

\noindent The experiments outlined below attempt to learn symbolic representations of image embeddings that are grounded in semantic visual concepts. These symbol sequence representations are learned through a paired-LSTM sender/receiver architecture (see Section~\ref{sec:emergent_language}) using two key loss formulations: i) a referential game loss (see Section~\ref{sec:referential}) adapted from previous work \cite{havrylov2017emergence}; and ii) a contrastive loss formulation emphasizing between-class discrimination (see Section~\ref{sec:contrastive}). Symbolic representations are classified using neural machine translation (NMT) and random forest classification (RFC); see Section~\ref{sec:classification}. Finally, two methods were utilized to better understand i) class-relevant visual evidence towards learned symbols (Gradient-weighted Class Activation Mappings; Grad-CAM); and ii) underlying embedding spaces (t-distributed Stochastic Neighbor Embedding; t-SNE) learned using contrastive methods (see Section~\ref{sec:visualization}). An overview of the systems utilized in each training paradigm is provided in Figure~\ref{fig:system_diagram}.

\subsection{Emergent Language Encoder-Decoder}
\label{sec:emergent_language}

\noindent The core architecture used to learn symbolic representations of image embeddings consists of two LSTM networks: a sender and receiver. The encoder (sender) network $Enc(\cdot)$ transforms embeddings $e$ into symbols $s_i$. This is achieved through categorical sampling of hidden layer activations $h_i^{enc}$ at each step $i$ ($1 \leq i \leq 10$; 10 words per sentence) in the LSTM: $s_i=C(\mathrm{softmax}(W h_i^{enc} + B))$. Note that an affine transformation $f(\cdot)$ was used to determine initial hidden layer activations: $h_0^{enc}=f(e)$. The decoder (receiver) network $Dec(\cdot)$ accepts symbols $s_i$ at each time step and returns reconstructed embeddings $e'$ at the final step by applying a second affine transformation $g(\cdot)$: $e'=g(h_{10}^{dec})$. Reconstructed embeddings are then compared with embeddings from either i) a set of distractor images as well as the true image in a referential game setup (see Section~\ref{sec:referential}; Figure~\ref{fig:system_diagram}, top) or ii) a positive (same category) and negative (different category) images in a contrastive loss formulation (see Section~\ref{sec:contrastive}; Figure~\ref{fig:system_diagram}, bottom).

In the referential game system, the sender and receiver LSTMs had two layers with a hidden size of 256; in the contrastive system, they had one layer with a hidden size of 256. Only one layer was needed in the contrastive system since the feature extractor network was trained alongside the encoder/decoder networks. The number of words in each sentence was 10 in each experiment, and vocabulary sizes of 1024 (based on prior work~\cite{havrylov2017emergence}) and 32 were used in the referential and contrastive cases, respectively. However, note that approximately 30 unique symbols were utilized following training in the referential case while all other symbols went unused. Regarding the dimensions of input embeddings, an embedding size of 1024 was used in the referential game setup based on output from a pretrained ResNet convolutional neural network (CNN). In the contrastive system, a novel CNN was trained end-to-end and consisted of three convolutional blocks with output dimensions of 64, 128, and 256. The resulting embedding size was therefore 256. The lower embedding size in the latter experiment was chosen due to the drastically smaller dataset used when training the convolutional layers ($N=400$ images over 20 categories compared to ImageNet's $N=50,000$ images over 1000 categories). 

\subsection{Referential Game}
\label{sec:referential}

\noindent One method for training two agents to develop a shared communication protocol is to have them play a referential game wherein a target image is selected and a sender agent generates a set of words to describe that image. The receiver image interprets those words and attempts to select the target image amongst a set of distractor images in a batch with 32 samples. While the agents' private language initially has no meaning, eventually the two agents reach common ground on how words relate to underlying image embeddings. In practice, this is achieved by comparing reconstructed embeddings $e'$ with their true embeddings $e$ using a hinge loss function. When selecting distractor images, class labels are not taken into account, so it is possible that other images belonging to the target class will be distinguished from the target itself. This forces the language to not only learn to discriminate between images of different classes but also images belonging to the same class. It is the opinion of the authors that this pushes the sender and receiver networks to learn compositional languages where additional words in a sentence provide further specification of semantics which can be useful for within-class discrimination; see~\cite{havrylov2017emergence}, Figure 3.

\subsection{Contrastive Learning} 
\label{sec:contrastive}

\noindent The contrastive loss system differs from the referential game system in two key ways: i) instead of matching the target (anchor) image's reconstructed embedding with the target's true embedding, it is matched with the embedding of a separate image belonging to the same class (positive example); and ii) distractor images (negative examples) are sampled from all other classes. The same hinge loss function is used to compare reconstructed (encoded and decoded) embeddings with true embeddings.

\subsection{Classification Methods}
\label{sec:classification}

\noindent In the referential game system, a random forest classification (RFC) model was used to predict class labels from symbols produced by the trained sender (encoder) network. This model was trained after the encoder and decoder networks converged on a learned communication protocol. However, in both the referential game and contrastive systems, a neural machine translation (NMT) network was utilized for class prediction. The purpose of implementing the RFC only in the referential game system was to compare performance with NMT methods--this comparison was not necessary in the contrastive case, as indicated by the results in Section~\ref{sec:experiments}. In terms of the NMT architecture, an embedding size of 64 was used along with two transformer encoder blocks. Each block consisted of two linear layers with dimension 256 and 64 followed by layer normalization and dropout layers. In additional, multi-head attention was used as the self attention mechanism.

\subsection{t-SNE and Grad-CAM Visualizations}
\label{sec:visualization}

\noindent In the contrastive learning system, a CNN feature extractor is trained alongside the sender/receiver LSTM networks. Since pretrained embeddings were not utilized in the contrastive system, the need for visualizing learned embeddings arises. To achieve this, high-dimensional embeddings were transformed to a lower-dimensional 2D space using the t-distributed Stochastic Neighbor Embedding (t-SNE) method \cite{van2008visualizing}. The purpose of this visualization is to ensure that images belonging to different classes form distinguishable clusters in this learned low-dimensional space. The t-SNE visualization also allows for assessment of whether images belonging to semantically similar classes, e.g., \emph{horse} and \emph{zebra}, belong to clusters that are closer to one another than those of semantically dissimilar classes.

While t-SNE visualizations provide us with insight on the learned embedding space, it remains unclear how symbols/words themselves relate to visual evidence in encoded images. To this end, we utilize Gradient-weighted Class Activation Mapping (Grad-CAM)~\cite{selvaraju2017grad} to localize important regions in images based on words in sentences and resulting predictions from the NMT classifier. The CNN feature extractor, encoder LSTM, and NMT predictor are frozen in this exercise, and NMT predictions are used to generate heat maps over pixels representing class-relevant pixel activations (on a word-by-word basis). This method also allows for the generation of counterfactual class activation mappings indicating the evidence in each image which distinguishes it from a (designated) separate class.

\subsection{Datasets}
\label{sec:datasets}

\begin{wrapfigure}{r}{0.5\textwidth}
  \vspace{-7mm}
  \begin{center}
    \includegraphics[width=0.48\textwidth]{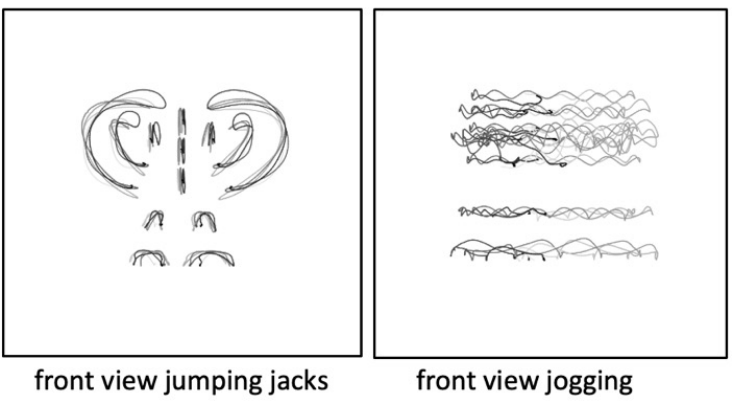}
  \end{center}
  \vspace{-3mm}
  \caption{Examples of two actions (jumping jacks, left; jogging, right) transformed into 2D images from 3D motion trajectory data in the MoVi dataset~\cite{ghorbani2021movi}.}
  \label{fig:actions}
  \vspace{2mm}
\end{wrapfigure}

\textbf{}

\vspace{-6mm} \noindent The aforementioned systems for symbolic encoding and class prediction (using neural machine translation and random forest classification) were applied to two datasets: the Sketchy dataset \cite{sangkloy2016sketchy} for object recognition (see Section~\ref{sec:exp_objects}) and the MoVi dataset \cite{ghorbani2021movi} for action recognition (see Section~\ref{sec:exp_actions}). The Sketchy dataset is comprised of drawings from human participants across 125 categories with approximately 100 examples (real pictures) of each; six sketches of each example were produced leading to approximately 600 images per class (see Figure~\ref{fig:system_diagram} for examples). The MoVi dataset is comprised of 3D trajectories of joint positions for 20 separate actions being performed over durations on the order of 10 seconds. Trajectory data was transformed to 2D contour images by drawing trajectories between each time step from a front-facing viewpoint with lower intensity corresponding with later time points (see Figure~\ref{fig:actions} for examples). For the Sketchy dataset, a set of 20 object classes were selected based on their semantic similarity; the chosen set consisted of classes which were both easy to distinguish (e.g., \emph{horse} and \emph{motorcycle}) and classes that were difficult to distinguish (e.g., \emph{horse} and \emph{zebra}). For the MoVi dataset, there are only 20 classes, each of which was used. A list of object and action categories can be found in the classification results in Tables \ref{tab:object_classification} and \ref{tab:action_classification}, respectively. A train/val/test split of 70/10/20 was used for both datasets.

\textbf{}

\begin{wraptable}{r}{.5\textwidth}
    \centering
    \vspace{16.8mm}
    \caption{Classification results for the objects dataset using referential game training (RFC-RG, NMT-RG) and contrastive learning training (NMT-CL).}
    \vspace{2mm}
    \begin{tabularx}{.5\textwidth}{l Y Y Y}
        \hline
        
        \textbf{Category} & \textbf{RFC-RG} & \textbf{NMT-RG} &  \textbf{NMT-CL} \\
        
        \hline

        ant & 40 & 43 & \textbf{89} \\
        axe & 27 & 25 & \textbf{79} \\
        butterfly & 17 & 18 & \textbf{93} \\
        car & 37 & 43 & \textbf{96} \\
        chair & 40 & 42 & \textbf{92} \\
        chicken & 20 & 21 & \textbf{80} \\
        couch & 48 & 41 & \textbf{94} \\
        fish & 12 & 12 & \textbf{55} \\
        hammer & 26 & 33 & \textbf{86} \\
        horse & 16 & 21 & \textbf{87} \\
        knife & 33 & 31 & \textbf{95} \\
        motorcycle & 41 & 41 & \textbf{94} \\
        owl & 15 & 13 & \textbf{87} \\
        seagull & 15 & 14 & \textbf{80} \\
        shark & 24 & 24 & \textbf{76} \\
        spoon & 46 & 51 & \textbf{95} \\
        table & 55 & 61 & \textbf{95} \\
        tree & 09 & 14 & \textbf{90} \\
        wheelchair & 46 & 59 & \textbf{89} \\
        zebra & 54 & 59 & \textbf{80} \\

        \hline
        
        \textbf{overall} & 31 & 33 & \textbf{88} \\

        \hline 
        
    \end{tabularx}
    \vspace{-23mm}
    \label{tab:object_classification}
\end{wraptable}

\textbf{}

\section{Experiments}
\label{sec:experiments}

\noindent The experiments in this section correspond with the datasets described in Section~\ref{sec:datasets}. For each dataset, a referential game training paradigm was used to learn symbolic representations of images using pretrained CNN (ResNet) embeddings. Symbols were subsequently classified using RFC and NMT prediction models. A contrastive learning paradigm was then explored wherein symbolic representations of embeddings from an initially untrained CNN were classified using a NMT prediction model. In the contrastive learning system, a t-SNE module mapped learned embeddings to a lower-dimensional space for visual exploration, and a Grad-CAM module visualized image regions associated with symbols used to classify objects and actions (using the NMT predictor).

\subsection{Experiment 1: Grounding Object Imagery}
\label{sec:exp_objects}

\noindent A referential game training paradigm was used to learn symbolic representations (sequences of integer symbols) of ResNet image embeddings. These symbols were used to classify images using RFC and NMT predictor models; RFC features were arrays of integer symbols concatenated with the frequency of each additional symbol. The decision to compare these models was based on the hypothesis that NMT methods are superior for classifying sequences of words since they capture the learned grammar implicit in the developed communication protocol. The RFC and NMT classifiers had validation accuracies of 31\% and 33\%, respectively, indicating superior classification performance by transformer-based methods in line with our hypothesis. We therefore utilized NMT classification for symbolic descriptions learned in the contrastive learning training paradigm which achieved a validation accuracy of 88\%. This significant improvement in accuracy is due to the increased degree of supervision during training. Classification accuracies for each predictor model, training paradigm, and object category are provided in Table~\ref{tab:object_classification}.

\begin{figure}[t]
  \includegraphics[width=\textwidth]{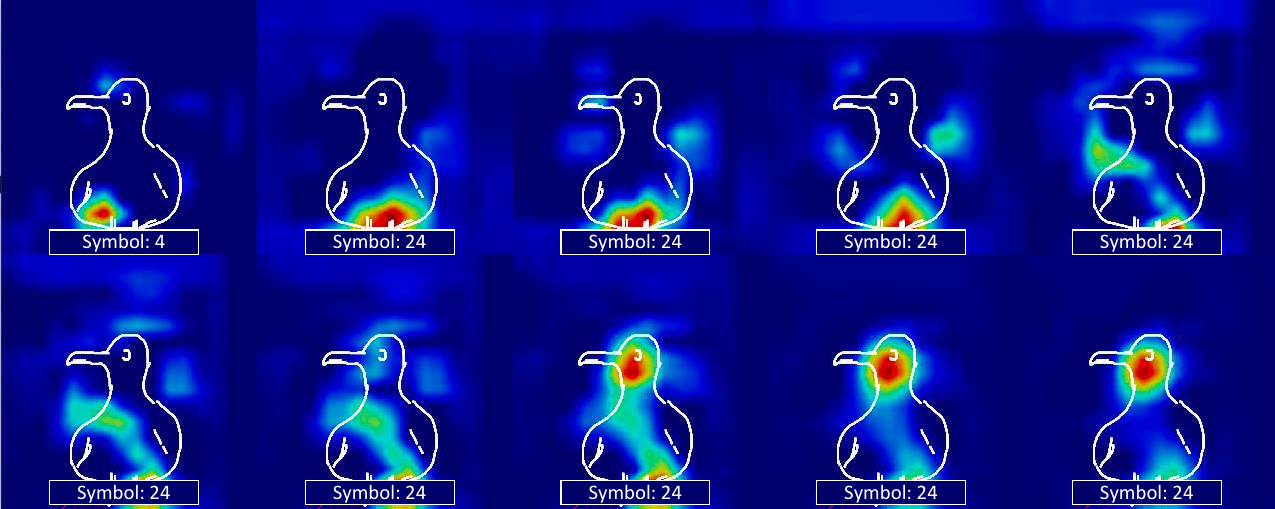}
  \caption{Grad-CAM visualizations for each of 10 symbols used to describe an image of a \emph{seagull}.}
  \label{fig:gradcam_seagull}
\end{figure}

We have shown thus far how symbolic representations of image embeddings can be used to predict object classes. Next, we aim to visualize evidence from object images as it relates to their symbolic descriptions. Specifically, we aim to determine whether visual evidence conveys semantic meaning which provides an explanation for why certain symbols are used to describe certain objects. To this end, a Grad-CAM module is used to compute gradients backwards from i) class labels predicted by the NMT model to ii) pixel activations used to generate words/symbols. An example of Grad-CAM results for an image of a \emph{seagull} is provided in Figure~\ref{fig:gradcam_seagull}. In this example, the sentence begins with the word (integer) 4 and the remainder of the sentence is comprised of the word (integer) 24 repeated nine times. One might assume that each instance of the word 24 corresponds to the same region in the image. However, as shown in the heat map, the word 24 is initially produced from pixels belonging to the seagull's legs and then shifts upwards, first to the seagull's body and then to its head. While the sentence used to describe this \emph{seagull} is not compositional in itself, the semantic meaning of the symbols appears to be so, i.e., the seagull concept is constructed by first conveying information about its feet, followed by its body and head.

\begin{wrapfigure}{r}{0.5\textwidth}
  \vspace{-2mm}
  \begin{center}
    \includegraphics[width=0.48\textwidth]{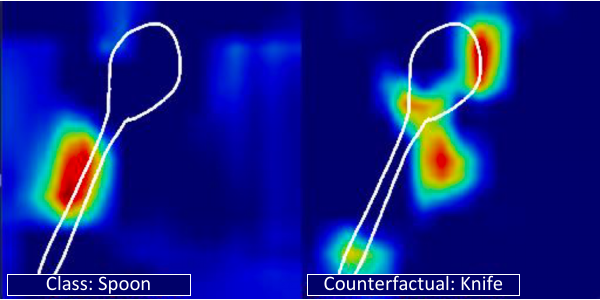}
  \end{center}
  \vspace{-1mm}
  \caption{Contrastive Grad-CAM from image of the \emph{spoon} class with counterfactual on the \emph{knife} class.}
  \label{fig:counterfactual_gradcam_spoon}
  \vspace{-5mm}
\end{wrapfigure}

In addition to visualizations of pixel evidence towards symbols with respect to image class, the Grad-CAM module allows for visualizations of evidence which differentiated the image from another class (i.e., counterfactual evidence). An example of evidence supporting correct classification of a \emph{spoon} along with evidence discriminating against a separate class \emph{knife} is provided in Figure~\ref{fig:counterfactual_gradcam_spoon}. In this case, the middle of the spoon's handle appears to provide positive evidence towards the \emph{spoon} label while the bowl of the spoon (and the bottom of the handle) appears to provide negative evidence towards the \emph{knife} label. While these results are based on the final word in the spoon's sentence, the same output can be generated for each word. This provides a method for not only understanding how a symbol captures what makes an object an object but also what makes that object different from other objects (with respect to the words \linebreak

\vspace{0mm} \noindent used to describe it). Taken together, these results demonstrate how explainable symbolic representations of objects can be constructed and how those representations can be tied into underlying visual semantics which define an object and discriminate it from others.

\begin{wrapfigure}{r}{0.5\textwidth}
  \vspace{-4mm}
  \begin{center}
    \includegraphics[width=0.48\textwidth]{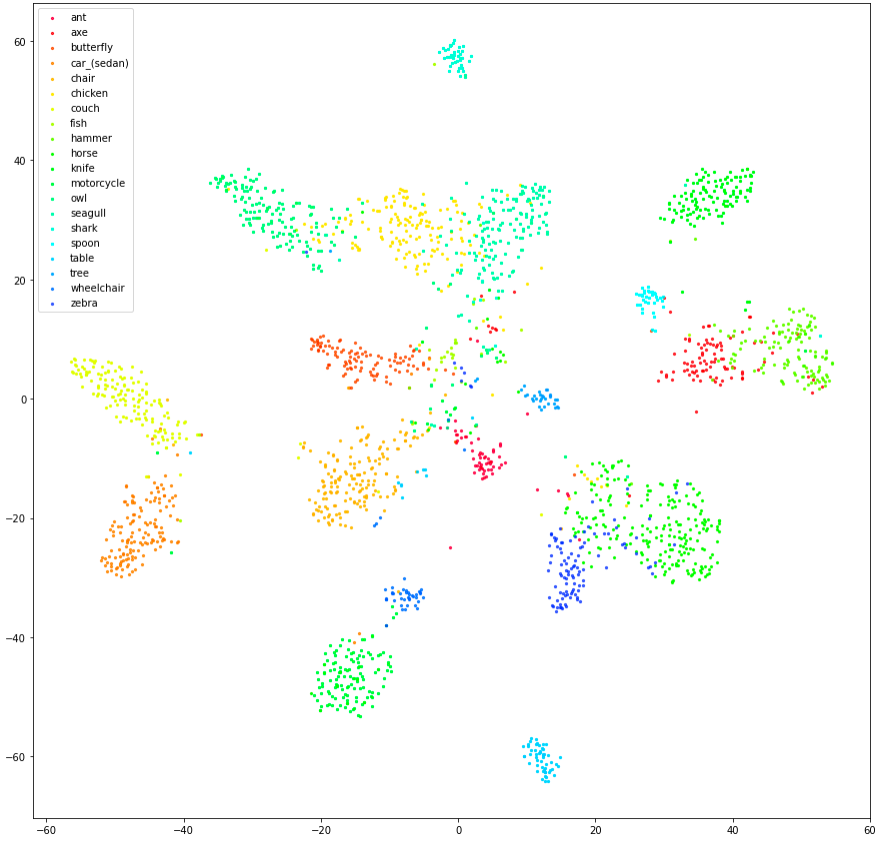}
  \end{center}
  \vspace{0mm}
  \caption{t-SNE plot for learned embeddings of object categories.}
  \label{fig:tsne_objects}
  \vspace{-10mm}
\end{wrapfigure}

The previous results show how symbolic representations can be related to visual semantic features in object imagery. Next, CNN embeddings from the trained feature extractor module were transformed to a lower-dimension space using the t-SNE method (see Figure~\ref{fig:tsne_objects}). Results show that images belonging to the same category appear to cluster together in the learned embedding space. Moreover, categories that are semantically similar appear to \linebreak

\vspace{-6mm} \noindent cluster in neighboring regions in the lower-dimensional space: e.g., horse {\color{horse} \newmoon} and zebra {\color{zebra} \newmoon} (bottom-right); chicken {\color{chicken} \newmoon}, owl {\color{owl} \newmoon}, and seagull {\color{seagull} \newmoon} (middle-top-left); and axe {\color{axe} \newmoon} and hammer {\color{hammer} \newmoon} (right). In contrast, categories which are semantically dissimilar appear to be clustered separately. This is consistent with confusion scores between classes (not included in this report).

\subsection{Experiment 2: Grounding Action Imagery}
\label{sec:exp_actions}

\noindent The same training paradigms and feature extraction methods outlined in the previous section were used to classify 2D images of action trajectories from 3D motion capture data. Consistent with previous results, the NMT prediction model following referential game training achieved a higher validation accuracy than the RFC prediction model; 55\% and 54\%, respectively. However, the difference in performance was not as notable as in the previous result. NMT classification performance was approximately the same for action categories as it was for object categories, achieving a validation accuracy of 87\%. Classification accuracies for each predictor model, training paradigm, and action category are provided in Table~\ref{tab:object_classification}.

\textbf{}

\begin{wraptable}{r}{.5\textwidth}
    \centering
    \vspace{-7mm}
    \caption{Classification results for the actions dataset using referential game training (RFC-RG, NMT-RG) and contrastive learning training (NMT-CL).}
    \vspace{2mm}
    \begin{tabularx}{.5\textwidth}{l Y Y Y}
        \hline
        
        \textbf{Category} & \textbf{RFC-RG} & \textbf{NMT-RG} &  \textbf{NMT-CL} \\
        
        \hline

        check watch & 38 & 39 & \textbf{87} \\
        crawl & 81 & 89 & \textbf{92} \\
        cross arms & 51 & 37 & \textbf{76} \\
        cross legs & 45 & 60 & \textbf{89} \\
        hand clap & 39 & 46 & \textbf{88} \\
        hand wave & 25 & 30 & \textbf{79} \\
        jog & 90 & 91 & \textbf{97} \\
        jumping jacks & 68 & 52 & \textbf{100} \\
        kicking & 42 & 47 & \textbf{90} \\
        phone talking & 17 & 37 & \textbf{83} \\
        point & 44 & 40 & \textbf{86} \\
        run in place & 85 & 66 & \textbf{94} \\
        scratch head & 52 & 57 & \textbf{85} \\
        sideways walk & 75 & 74 & \textbf{100} \\
        sit down & 57 & 72 & \textbf{88} \\
        stretch & 19 & 49 & \textbf{71} \\
        take photo & 32 & 35 & \textbf{76} \\
        throw catch & 39 & 58 & \textbf{73} \\
        vertical jump & 60 & 46 & \textbf{94} \\
        walk & 88 & 83 & \textbf{92} \\

        \hline
        
        \textbf{overall} & 54 & 55 & \textbf{87} \\

        \hline 
        
    \end{tabularx}
    \vspace{-4mm}
    \label{tab:action_classification}
\end{wraptable}

\textbf{}

\begin{figure}[b]
  \vspace{-6mm}
  \includegraphics[width=\textwidth]{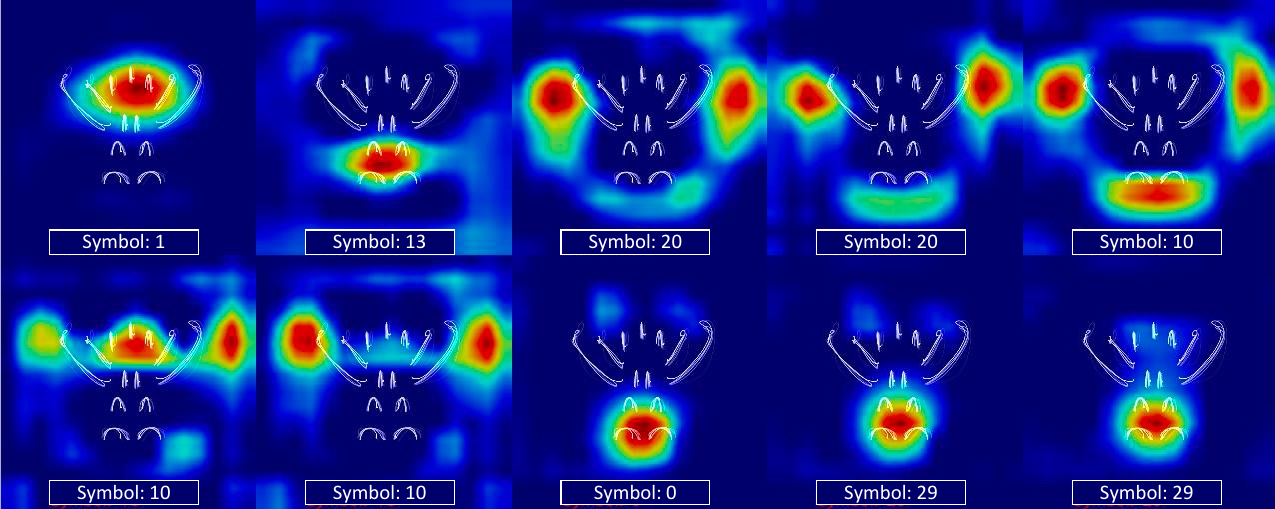}
  \caption{Grad-CAM visualizations for each of 10 symbols used to describe an image of \emph{jumping jacks}.}
  \label{fig:gradcam_jumping_jacks}
\end{figure}

Next, we examine Grad-CAM results for an image of a participant performing \emph{jumping jacks} (see Figure~\ref{fig:gradcam_jumping_jacks}). In contrast with the previous experiment, the image is described by multiple unique integer symbols: 0, 1, 10, 13, 20, and 29. Observe that i) the symbol 1 appears to capture the motion of the torso, ii) the symbol 13 captures the feet and knees, iii) the symbol 20 places additional attention on the feet, iv) the symbol 10 captures the feet and then moves to include the torso, and v) the remainder of the symbols focus on the feet and knees. As in the first experiment, the language appears to describe imagery by conveying information in a piecemeal fashion; the language in this example, however, demonstrates a higher degree of compositionality. We hypothesize that this increased level of compositionality is needed to capture this dataset's inherent within-class variability as mentioned above. It is important to note here that the same symbol does not always mean the same thing; symbols are used in different ways for different object categories. This means that, for example, the symbol 1 does not always point to the torso, and the symbol 13 does not always point to the legs and knees. While in natural language the same word can mean different things, the learned languages described herein take this notion to the extreme.

\begin{wrapfigure}{r}{0.475\textwidth}
  \vspace{-6mm}
  \begin{center}
    \includegraphics[width=0.475\textwidth]{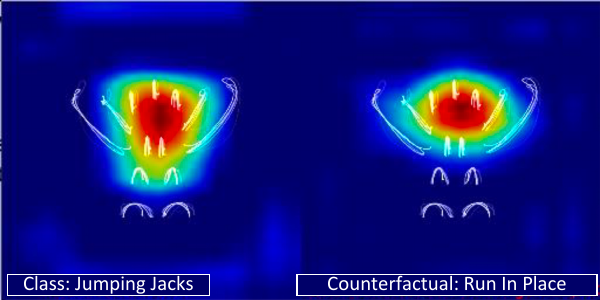}
  \end{center}
  \vspace{-2mm}
  \caption{Contrastive Grad-CAM from image of the \emph{jumping jacks} class with counterfactual on the \emph{run in place} class.}
  \label{fig:counterfactual_gradcam_jumping_jacks}
  \vspace{8mm}
\end{wrapfigure}

We conclude by providing an example of evidence supporting correct classification of \emph{jumping jacks} along with evidence discriminating against a separate class \emph{run in place} (see Figure~\ref{fig:counterfactual_gradcam_jumping_jacks}). In this case, the participant's head, shoulders, hips, and knees appear to provide positive evidence towards the \emph{jumping jacks} label while the elbows and hands appear to provide negative evidence towards the \emph{run in place} label (note the major axis of the activation ellipsoid is vertical for \emph{jumping jacks} and horizontal for \emph{run in place}).

\begin{wrapfigure}{r}{0.5\textwidth}
  \vspace{-65mm}
  \begin{center}
    \includegraphics[width=0.48\textwidth]{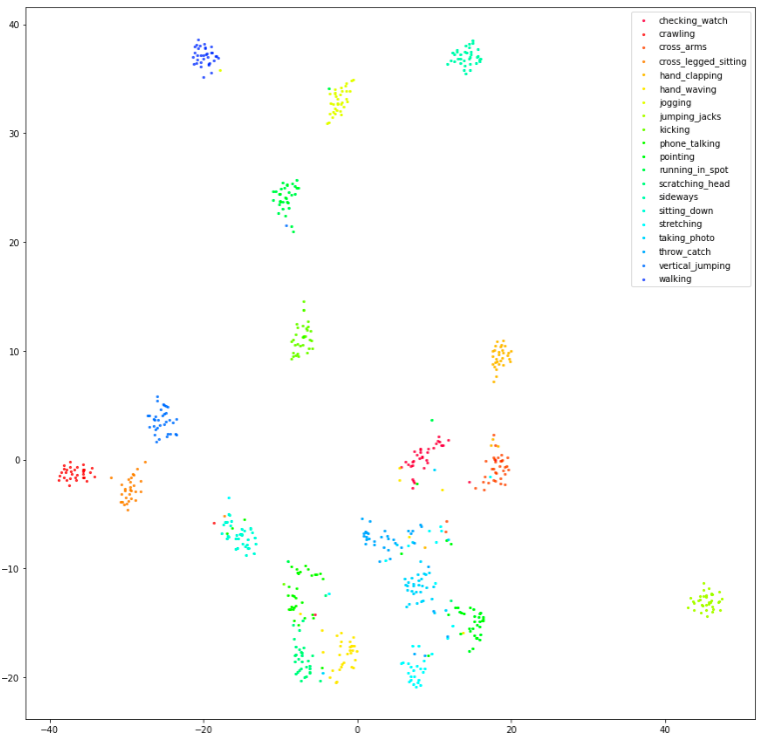}
  \end{center}
  \vspace{-1mm}
  \caption{t-SNE plot for learned embeddings of action categories.}
  \label{fig:tsne_actions}
  \vspace{0mm}
\end{wrapfigure}

Visualizations of learned t-SNE representations for action categories are provided in Figure~\ref{fig:tsne_actions}. As before, images belonging to the same category appear to cluster together and categories that are semantically similar cluster together in neighboring regions: e.g., hand wave {\color{hand_wave} \newmoon} and scratch head {\color{scratch_head} \newmoon} (bottom-bottom-left); and take photo {\color{take_photo} \newmoon} and throw catch {\color{throw_catch} \newmoon} (bottom-center). However, the degree of cluster proximity was not as apparent in the action dataset compared with the \linebreak

\vspace{-5.5mm} \noindent object dataset. While confusion between object categories was generally proportional to the nearness of their clusters, confusion between action categories generally occurs when only a few examples of a category are embedded in regions far from their class members into other class' clusters. This may be due to the large degree of variation in which participants performed their assigned actions when being recorded.

\vspace{-1mm}
\section{Conclusion}
\label{sec:conclusion}

\noindent Our experiments demonstrate how private languages can be learned using two training methodologies (referential and contrastive) and how those symbols can be i) used to make class predictions and ii) related to visual semantic features at the pixel level. This serves as only the first step in transforming visual evidence from sensory modalities in the world into higher-level conceptual structures which can be reasoned over and generalized to novel situations. While the words used in learned languages did not hold consistent meaning across image categories, we aim to examine and enable this capability (i.e., the unsupervised disentanglement of concepts) in future work. Given that symbolic representations of semantic visual features can be generated, an Agent must then learn how those features relate to the underlying attributes and affordances that define (and describe) an entity--and what those characteristics mean with respect to the entity's role in the world. We aim to pursue this research direction in future work by exploring the process of discovering concepts and their relations and placing those discoveries in the context of symbol manipulation in grounded languages.

\bibliographystyle{unsrt}  
\bibliography{arXiv_ALAS_contrastive}  






\end{document}